\pdfoutput=1
\documentclass{article}
\usepackage{spconf,amsmath,amssymb,amsthm,graphicx,booktabs,makecell,url}
\usepackage[hidelinks]{hyperref}
\usepackage{xcolor}
\usepackage{eso-pic}
\AddToShipoutPictureFG*{%
  \AtPageLowerLeft{%
    \put(\LenToUnit{0.5\paperwidth},\LenToUnit{24pt}){%
      \makebox[0pt][c]{\parbox{0.62\paperwidth}{\centering\scriptsize
        This work has been submitted to the IEEE for possible publication.
        Copyright may be transferred without notice, after which this version
        may no longer be accessible.}}}}}

\usepackage[labelsep=period,labelfont=bf,skip=10pt]{caption}
\usepackage{subcaption}
\def\x{{\mathbf x}}
\newcommand{\R}{\mathbb{R}}
\newcommand{\supp}{\mathcal{S}}
\newcommand{\jasin}{\mathrm{JaSIN}}
\newcommand{\relu}{\mathrm{ReLU}}
\IfFileExists{tab/macros.tex}{\newcommand{\ladderbounded}{0.06}
\newcommand{\ladderbump}{3.29}
\newcommand{\laddercentred}{3.64}
\newcommand{\laddersupport}{6.93}
\newcommand{\ladderosc}{6.30}
\newcommand{\laddertotal}{13.30}
\newcommand{\ladderrelu}{24.0}
\newcommand{\ladderjasin}{37.3}

\newcommand{\deadladdertri}{58}
\newcommand{\deadladderone}{5}
\newcommand{\ladderimages}{24}
\newcommand{\ladderseeds}{3}
\newcommand{\prunemaxparams}{84}
\newcommand{\prunemaxspeed}{3.2}
\newcommand{\prunecostmin}{1.1}
\newcommand{\prunecostmax}{11.9}
\newcommand{\pruneworstunits}{1.5}
\newcommand{\prunebestunits}{57.8}

\newcommand{\certoverheadmin}{15}
\newcommand{\certoverheadmax}{35}
\newcommand{\cpuspeedup}{7.5}
\newcommand{\cpuspeedupwire}{17}

}{}

\IfFileExists{tab/macros.tex}{%
  \let\oldbounded\ladderbounded \renewcommand{\ladderbounded}{\oldbounded}%
  \let\oldsupport\laddersupport \renewcommand{\laddersupport}{\oldsupport}%
  \let\oldosc\ladderosc         \renewcommand{\ladderosc}{\oldosc}%
}{}
\IfFileExists{fig/qualitative_labels.tex}{\newcommand{\qualpsnrrelu}{17.7}
\newcommand{\qualpsnrjasin}{32.1}
\newcommand{\qualpsnrsiren}{33.8}
\newcommand{\qualimage}{kodim08}
}{}
\IfFileExists{fig/which_labels.tex}{\newcommand{\whichsupa}{13.1}
\newcommand{\whichosca}{5.0}
\newcommand{\whichsupb}{9.5}
\newcommand{\whichoscb}{4.8}
\newcommand{\whichsupc}{5.8}
\newcommand{\whichoscc}{6.1}
\newcommand{\whichsupd}{4.6}
\newcommand{\whichoscd}{9.0}
}{}

\providecommand{\certoverheadmin}{14}
\providecommand{\certoverheadmax}{35}
\providecommand{\cpuspeedup}{7.5}
\providecommand{\cpuspeedupwire}{17}
\providecommand{\qualimage}{kodim08}
\providecommand{\qualpsnrrelu}{??}\providecommand{\qualpsnrjasin}{??}
\providecommand{\qualpsnrsiren}{??}
\providecommand{\whichsupa}{??}\providecommand{\whichosca}{??}
\providecommand{\whichsupb}{??}\providecommand{\whichoscb}{??}
\providecommand{\whichsupc}{??}\providecommand{\whichoscc}{??}
\providecommand{\whichsupd}{??}\providecommand{\whichoscd}{??}
\renewcommand{\paragraph}[1]{\par\smallskip\noindent\textbf{#1.}}
\providecommand{\tabsep}{3pt}
\newtheorem{proposition}{Proposition}

\providecommand{\ladderbounded}{0.05}
\providecommand{\ladderbump}{3.31}
\providecommand{\laddercentred}{3.46}
\providecommand{\ladderosc}{6.62}
\providecommand{\laddertotal}{13.45}
\providecommand{\ladderrelu}{24.1}
\providecommand{\ladderjasin}{37.6}

\providecommand{\deadladdertri}{58}
\providecommand{\deadladderone}{5}
\providecommand{\ladderimages}{24}
\providecommand{\ladderseeds}{3}
\providecommand{\auditevery}{200}
\providecommand{\audititers}{5000}
\providecommand{\auditruns}{18}
\providecommand{\revivalhigh}{0.9}
\providecommand{\revivalmid}{56}
\providecommand{\revivallow}{3.1}
\providecommand{\prunemaxparams}{84}
\providecommand{\prunemaxspeed}{3.2}
\providecommand{\prunecostmin}{1.1}
\providecommand{\prunecostmax}{11.9}
\providecommand{\pruneworstunits}{1.5}
\providecommand{\prunebestunits}{57.8}

\title{A Piecewise-Linear Activation Function for Implicit Neural Representations}
\name{Shahd Hekal, Joram Kim, Jeova Farias Sales Rocha Neto}
\address{Bowdoin College, Brunswick, ME, USA}

\begin{document}
\ninept
\maketitle

\begin{abstract}
Coordinate networks with ReLU activations perform poorly on implicit neural representation tasks, and the activations that address this, such as sinusoids and Gabor wavelets, all leave the piecewise-linear function class, so it remains unclear which property of ReLU is at fault. We introduce JaSIN, a family of bounded piecewise-linear activations that interpolates between ReLU and a piecewise-linear sinusoid without any the use of transcendental operation. Our results show that a bounded range improves image reconstruction PSNR on average by 
$\ladderbounded$~dB, a single bump $\ladderbump$~dB, and centering that bump a further $\laddercentred$~dB. The oscillation count
improves $\ladderosc$~dB more, raising ReLU from $\ladderrelu$ to
$\ladderjasin$~dB on Kodak images, placing third over eleven activations, within 1.3 dB of the state-of-the-art, though beyond four oscillations the reconstruction collapses. Because an INR is fit on the grid on which it is evaluated, a unit whose pre-activations never reach the support
of its activation is certifiably inactive over the entire domain; deleting such
units is lossless without fine-tuning, removing up to $\prunemaxparams\%$ of parameters at
zero loss in performance (dB) and decoding $\prunemaxspeed\times$ faster.
Reference code and demos are available at
\url{https://github.com/jeovafarias/jasin}.
\end{abstract}


\section{Introduction}
\label{sec:intro}
An Implicit Neural Representation (INR) encodes a signal as a small coordinate
network $f_\theta:\R^d\!\to\!\R^c$ fit to samples on a grid \cite{essakine2024we}. Coordinate networks
with ReLU activations perform poorly at this task, and the field has responded
by replacing the nonlinearity with a sinusoid, a wavelet or a bump. Every one of
those alternatives leaves the piecewise-linear function class and introduces a
transcendental operation at the same time, so the diagnosis is never isolated.
It is known that ReLU is insufficient, but not whether the missing property is
a bounded range, two-sided support, smoothness or oscillation.

In this work we introduce JaSIN, a family of piecewise-linear activations that
interpolates between ReLU and a piecewise-linear sinusoid, and we use it to
identify which property of ReLU is lacking when applied to INRs. Beyond that, we also provide a few more complimentary results. Whenever an activation vanishes on a known set, a unit whose
pre-activations never leave that set can be certified inactive. For a compactly
supported activation the test becomes two-sided, and because an INR is fit on the grid on which it is evaluated, the certificate is
exact. We use it to repair networks during training and to
compress them afterwards, at no loss and without retraining.

\section{Related work and contributions}
\label{sec:related}
\vspace{-4pt}
\paragraph{Activations for implicit representations}
The remedies for ReLU's poor high-frequency capacity are all transcendental:
random Fourier features precede the network with a fixed sinusoidal
embedding~\cite{tancik2020fourier}, SIREN makes the activation itself a
sinusoid~\cite{sitzmann2020siren}, WIRE uses a Gabor
wavelet~\cite{saragadam2023wire}, and Gaussian~\cite{ramasinghe2022gaussian} and
variable-periodic~\cite{liu2024finer} activations vary the bandwidth. Each
supplies oscillation, a bounded range and smoothness together, which is why the
literature reports that ReLU is insufficient without saying what it lacks.

\paragraph{Piecewise-linear activations}
These are well catalogued outside the INR setting. The tent activation
$\min(\relu(1{-}z),\relu(1{+}z))$ and the leaky single-peaked triangle
unit~\cite{shan2020lsptlu} have been proposed
repeatedly~\cite{kunc2024three}, and learnable families subsume
them~\cite{agostinelli2015learning,tavakoli2021splash}; all are learned,
initialized near ReLU and evaluated on classification, and none has been applied
to implicit representations. Oscillatory activations have been studied
separately, with some works claiming that certain types of oscillation improve classification
accuracy~\cite{noel2021oscillating}. Finally, the link between oscillation count and
capacity is the seminal depth-separation construction of
Telgarsky~\cite{telgarsky2016benefits}.

\paragraph{Unused units}
That a ReLU unit can stop firing permanently is the dying-ReLU
phenomenon~\cite{lu2019dying}, and counting near-zero activations over a sample
is the standard data-driven pruning criterion~\cite{hu2016network}. Making it
exact requires certifying that a unit is inactive everywhere, which for ReLU is
the stably inactive condition used for lossless
compression~\cite{serra2020lossless,serra2021scaling} and as a presolve step in
verification~\cite{tjeng2019evaluating,zhang2018crown}; those methods bound
pre-activations over a continuous region because the test inputs are unobserved.
Repair by reinitialization has also been proposed
before~\cite{qiao2019neural,sokar2023dormant}.

\paragraph{Contributions}
Broadly, what is novel, to the best of our knowledge, is the family that isolates
the diagnosis for the low performance of piece-wise linear function in INR research, and the setting in which the inactivity test is exact rather
than heuristic: implicit representations. Our specific
contributions are:
\begin{itemize}\setlength{\itemsep}{0pt}\setlength{\parskip}{0pt}
\setlength{\topsep}{2pt}
\item \textbf{A piecewise-linear activation function family} with an oscillation count, swept one property at
a time, which shows ReLU to be missing two properties that leads to its underperformance in INRs, and whose best member places third of eleven activations,
within $1.3$~dB of a tuned SIREN and ahead of every non-sinusoidal baseline on
every image, all of that using no transcendental operation;
\item \textbf{An exact inactivity certificate}: compact two-sided support turns the ReLU
stably inactive condition into an interval test, and an INR is fit on its whole
input domain, so the interval is computed rather than bounded;
\item \textbf{Lossless INR network compression without retraining}, removing up to $\prunemaxparams\%$ of
the parameters at $0.000$~dB and decoding $\prunemaxspeed\times$ faster, where
magnitude pruning at the same sparsity costs up to $\prunecostmax$~dB.
\end{itemize}

\section{Methodology}
\label{sec:method}

\subsection{The JaSIN family}
\label{sec:family}
\begin{figure}[t]
\centering
\includegraphics[width=\linewidth]{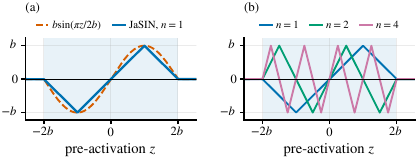}
\caption{The JaSIN family. (a) At $n=1$, a windowed triangle wave: the
piecewise-linear interpolant of a sinusoid. (b) The oscillation count $n$
varies at fixed support.}
\label{fig:family}
\end{figure}

In this work, we propose JaSIN (for \textbf{Ja}gged \textbf{SIN}e), a piecewise-linear approximation of a windowed
sinusoid. Let $T$ denote the unit triangle wave of period $4$, and let $b>0$ be
a half-width and $n\ge1$ an oscillation count. We define
\begin{equation}
\label{eq:family}
\jasin_{n}(z;b)=
\begin{cases}
 b\,T(nz/b), & |z| < 2b,\\
 0, & \text{otherwise,}
\end{cases}
\end{equation}
which is odd, has range $[-b,b]$, slope $\pm n$, and support
$\supp_b=(-2b,2b)$. At $n=1$, $\jasin$ is one period of a triangle wave under a
rectangular window, the piecewise-linear interpolant of $b\sin(\pi z/2b)$ on
its knots (Fig.~\ref{fig:family}a). Increasing $n$ packs
more oscillations into $\supp_b$ (Fig.~\ref{fig:family}b), and removing
the window recovers a periodic triangle wave, the piecewise-linear analogue of
$\sin(\omega_0 z)$.

Every member is piecewise linear, with $2n$ interior knots at the odd multiples
of $b/n$ inside $\supp_b$ and two at its edges, so it is an exact finite sum of
$2n+2$ rectifiers. We build it from the one-sided triangle
$\Lambda(z)=\min(\relu(z),\relu(2b{-}z))$ of Section \ref{sec:exp} where
$\min(p,q)=p-\relu(p{-}q)$, which at $n=1$ gives
\begin{equation}
\label{eq:relu}
\begin{aligned}
\jasin_1(z;b)&=\Lambda(z)-\Lambda(-z)\\
&=\relu(z{-}2b)-2\relu(z{-}b)\\[-2pt]
&\quad{}+2\relu(z{+}b)-\relu(z{+}2b),
\end{aligned}
\end{equation}
one rectifier per knot and none elsewhere. JaSIN is thus not a new
arithmetic primitive but an arrangement of the one ReLU networks already use,
which lets Section \ref{sec:exp} vary the shape of the nonlinearity at fixed
arithmetic class. An important property of Eq. \ref{eq:family} is that,operationally, it does not depend on the evaluation of a transcendent function, which is typical in INR  \cite{essakine2024we}.

\subsection{Properties}
\label{sec:props}

\paragraph{Gauge freedom}
Every member of the family is positively homogeneous:
$\jasin_n(z;\lambda b)=\lambda\,\jasin_n(z/\lambda;b)$. Hence $b$ can be
absorbed into the neighboring linear layers by rescaling $W^{(1)}$, the biases
and the output layer, a symmetry such activations are known to
admit~\cite{neyshabur2015path}. Consequently $b$ is not a hyperparameter of
the function class: varying it at fixed initialization varies the ratio of
support width to pre-activation spread.

\paragraph{Inactive units}
Let $X$ be the training set,
$z_j(\x)=\langle w_j,h^{(l-1)}(\x)\rangle+\beta_j$ the pre-activation of unit
$j$ at $\x$, and $I_j=[\min_X z_j,\allowbreak\;\max_X z_j]$ the interval it
sweeps. Call a unit \emph{inactive} on $X$ if its output is zero at every training
point.

\newpage
\begin{proposition}
\label{prop:cert}
Suppose $I_j\cap\supp_b=\emptyset$. Then
$\jasin_n(z_j(\x))=0$ for every $\x\in X$, and the gradients of the loss with
respect to $w_j$ and $\beta_j$ are exactly zero: unit $j$ receives no gradient of
its own, and under gradient descent, where its incoming features are also fixed
as in the first layer, the condition persists.
\end{proposition}
\vspace{-6pt}
\begin{proof}
Both $\jasin_n$ and $\jasin_n'$ vanish outside $\supp_b$, so the output and the
local gradient are zero on $X$; a gradient step leaves $w_j$ and $\beta_j$ where
they were, and if $h^{(l-1)}$ is unchanged as well then $I_j$ is unchanged and
the hypothesis holds at the next step.
\end{proof}
\vspace{-5pt}
Two things weaken that conclusion in practice. Adam updates $w_j$ and $\beta_j$
from moments accumulated before the gradient vanished, so an inactive unit
continues to move until those decay, and, beyond the first layer $h^{(l-1)}$
moves with the layers before it. The inactivity certificate above thus describes the current
parameters rather than the rest of the training process, and we empirically measure the difference. We
recompute it over the whole signal every $\auditevery$ steps of
$\audititers$, in $\auditruns$ runs with repair (defined below) disabled, the fraction of
certified units active again at a later audit is $\revivalhigh\%$ at $b=0.1$,
$\revivalmid\%$ at $b=0.05$ and $\revivallow\%$ at $b=0.02$, the setting at
which repair matters most. No first-layer unit ever revives in any run.
The inactivity certificate is therefore recomputed wherever it is used: repair
reads it during training, compression once at convergence; after which no
parameter changes and the certified unit set is final.

For ReLU the same test is the stably inactive condition of the compression and
verification literatures~\cite{serra2020lossless,tjeng2019evaluating}, with a
half-line in place of an interval; there the property must be decided over a
continuous region of unobserved inputs, which is NP-hard. An INR is fit on its
whole domain, so $I_j$ is computed rather than bounded: a running minimum and
maximum over the pre-activations the ordinary forward pass already produces, at
two values per unit and no extra pass. The extrema have to be accumulated over
the whole domain: on a subset $X'\subset X$ the interval is narrower, so a unit
that is active elsewhere can be certified in error. Empirically, we find that monitoring them at every training step costs an extra
$\certoverheadmin$--$\certoverheadmax\%$ compute against the fastest
implementation of each activation. Where only the final certificate
is needed, one pass at convergence suffices. Globally supported activations
admit no such test at all. A brute-force scan over the training set agrees with
the criterion exactly, on every activation we test.

\paragraph{Repair}
Every $K$ iterations we redraw the incoming weights of inactive units from the
layer's initialization law $U(-1/\sqrt{n_{\mathrm{in}}},1/\sqrt{n_{\mathrm{in}}})$,
for $n_{\mathrm{in}}=n_{l-1}=\dim h^{(l-1)}$ the fan-in of the layer repaired,
and set their biases so the mean pre-activation over a probe batch lands on the
midpoint of its active region, the origin since every member is odd. A unit
restarted there sits at the center of a linear piece of width $2b/n$, so its
pre-activation must travel $b/n$ to reach a kink, whereas the same construction
on a one-sided support places it at a kink immediately. We reset the unit's
optimizer moments, since stale momentum otherwise returns it to inactivity, and
touch only certified units, so a network with none is left exactly unchanged.
The naive baseline redraws the same weights from $0.01\,\mathcal{N}(0,1)$ and
zeroes the bias, restoring a gradient path but leaving the pre-activation at an
uncontrolled distance from $\supp_b$.


\paragraph{Compression}
An inactive unit contributes exactly zero, so deleting it and rebuilding with
narrower layers leaves the function unchanged. Deleting one leaves every other
unit's pre-activation untouched, its contribution having been zero already, so
the certified set is removed in one pass. Here the training set is the
entire input domain, so the guarantee is complete and no retraining is needed.
Deletion is structured, so the reduction in compute is achieved with standard dense arithmetic.

\section{Experiments}
\label{sec:exp}

\subsection{Experimental setup}
\label{sec:setup}
Kodak~\cite{kodak} is the standard reference set for image compression and
restoration: $24$ uncompressed photographs of $393{,}216$ pixels in three
channels, eighteen landscape and six portrait. We fit every image at its native
resolution, without cropping, resizing or conversion to grayscale, so a
$5\!\times\!256$ network holds $0.22$ parameters per target.

All activations share one architecture, $5$ hidden layers of $256$ units, the
SIREN reference configuration~\cite{sitzmann2020siren}, and one optimizer,
Adam~\cite{kingma2015adam} with batch size $8192$, $5000$ iterations and cosine decay. Where repair is
applied it runs every $K\!=\!50$ iterations. The sinusoidal baselines keep the
initialization their authors specify and every member of our family keeps the
Kaiming-uniform default~\cite{he2015delving}, so each baseline is given the
scheme it was designed for and the $\jasin$ family is shown to need none of its own.
Learning rates and shape parameters ($\omega_0$, $s_0$, $\sigma$, $b$) are
selected at this resolution and budget, every grid extended until the selected
value was interior. Empirically, we find $b\!=\!0.1$ to be optimal for $n\!=\!4$ on all $24$
images.

We report Peak signal-to-noise ratio (PSNR) and
Structural Similarity Index Measure (SSIM) ~\cite{hore2010image,wang2004image} over all pixels, with signals mapped to
$[-1,1]$. Every configuration is run with $3$ seeds on all $24$ images and we report the
mean results (PSNR and SSIM) over those $72$ runs with their standard deviation. The spread found in the following results mainly measures
image difficulty, not seed noise: within an image the seed-to-seed deviation is
$0.12$--$0.68$~dB against $2.7$--$3.5$~dB across images. We also report the effective rank of each layer's activations, i.e. the exponential of
the entropy of its normalized singular-value
spectrum~\cite{roy2007effective}, which counts the directions a layer actually
uses. Of the $256$ directions available due the tested layer dimension, SIREN reaches a $241$ effective rank and ReLU $39$. All runs use PyTorch $2.12$ on one RTX~4070 Laptop GPU and the CPU timings of Section \ref{sec:results} use one thread of an Intel i9-13900H.

\subsection{Results}
\label{sec:results}

\begin{table}[t]
\centering\scriptsize
\caption{From ReLU to a sinusoid, one property at a time, on $\ladderimages$
Kodak images with $\ladderseeds$ seeds. Each row adds the property in the
second column to the row above it. All rows are piecewise linear except
SIREN, which is included for reference. Each row is run at its own
selected learning rate. A dash (``--'') marks an inactive fraction that
is undefined for an activation with unbounded support.}
\label{tab:ladder}
\vspace{-5pt}
{\providecommand{\tabsep}{3pt}\setlength{\tabcolsep}{\tabsep}
\begin{tabular*}{\linewidth}{@{\extracolsep{\fill}}llccc@{}}
\toprule
Activation & adds & PSNR (dB) & inactive (\%) & effective rank \\
\midrule
ReLU & -- & 23.98\,\tiny{$\pm$2.98} & 41.6 & 39 \\
Clipped ReLU & bounded range & 24.04\,\tiny{$\pm$3.00} & 42.0 & 41 \\
\midrule
Triangle, 1-sided & one bump & 27.33\,\tiny{$\pm$3.16} & 57.8 & 79 \\
\textit{JaSIN, $n{=}1$} & two-sided & 30.97\,\tiny{$\pm$3.52} & 4.5 & 184 \\
\midrule
\textit{JaSIN, $n{=}2$} & 2 osc. & 34.59\,\tiny{$\pm$3.01} & 1.8 & 226 \\
\textit{JaSIN, $n{=}4$} & 4 osc. & 37.28\,\tiny{$\pm$2.99} & 1.5 & 238 \\
\textit{JaSIN, $n{=}8$} & 8 osc. & 15.19\,\tiny{$\pm$2.55} & 1.3 & 247 \\
\textit{JaSIN, $n{=}16$} & 16 osc. & 14.61\,\tiny{$\pm$2.12} & 1.4 & 248 \\
periodic tri. & no window & 34.50\,\tiny{$\pm$3.04} & --- & 222 \\
SIREN & -- & 38.55\,\tiny{$\pm$2.86} & --- & 241 \\
\bottomrule
\end{tabular*}
}
\end{table}

\paragraph{Ablation Study} Table~\ref{tab:ladder} sweeps the $\jasin$
family one property at a time. Clipping ReLU at a ceiling, $\min(\relu(z),1)$,
bounds its range but not its support and contributes $\ladderbounded$~dB, which
is negligible. Introducing a single bump contributes $\ladderbump$~dB, and centering
that bump on the origin contributes a further $\laddercentred$~dB while reducing
the inactive fraction from $\deadladdertri\%$ to $\deadladderone\%$. The latter follows from the geometry of the
support, since pre-activations concentrate near zero at initialization, which
lies in the interior of a two-sided support but on the boundary of a one-sided
one. The oscillation count contributes $\ladderosc$~dB, and the complete path from
ReLU to $n\!=\!4$ accounts for $\laddertotal$~dB. The gain is not monotone in
$n$: at $n\!=\!8$ and beyond reconstruction collapses to about $15$~dB even as
the effective rank keeps rising, so $n\!=\!4$ is the operating point we carry
forward and Section \ref{sec:disc} returns to the collapse.

\vfill

\begin{figure}[t]
\centering
\begin{subfigure}{0.242\linewidth}\centering
  \includegraphics[width=\linewidth]{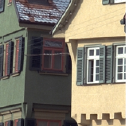}
  \caption{ground truth\\\vphantom{$0$~dB}}\end{subfigure}\hfill
\begin{subfigure}{0.242\linewidth}\centering
  \includegraphics[width=\linewidth]{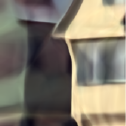}
  \caption{ReLU\\($\qualpsnrrelu$~dB)}\end{subfigure}\hfill
\begin{subfigure}{0.242\linewidth}\centering
  \includegraphics[width=\linewidth]{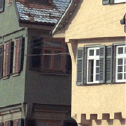}
  \caption{JaSIN, $n{=}4$\\($\qualpsnrjasin$~dB)}\end{subfigure}\hfill
\begin{subfigure}{0.242\linewidth}\centering
  \includegraphics[width=\linewidth]{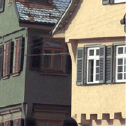}
  \caption{SIREN\\($\qualpsnrsiren$~dB)}\end{subfigure}
\vspace{-5pt}
\caption{A $200\!\times\!200$ crop of image \qualimage{} from the Kodak dataset at native resolution. The depicted PSNR for each activation is
computed over the full image.}
\label{fig:qual}
\end{figure}

\paragraph{Tuned baselines} Table~\ref{tab:native} places JaSIN
among established activations. The best member, $n\!=\!4$, ranks third of
eleven, $1.00$~dB below FINER and $1.27$~dB below SIREN, both significant under
a paired test ($p<10^{-7}$). It
reaches this range without any transcendental operation, however, and leaves
only $1.5\%$ of its units inactive against $16.1\%$ for ReLU with random
Fourier features. Figure~\ref{fig:qual} shows a crop of kodim08, the most difficult image in the
set. ReLU resolves neither the window frames nor the roof tiles, $n\!=\!4$
recovers both, and what separates it from SIREN is the highest-frequency detail.

\begin{table}[t]
\centering\scriptsize
\caption{Reconstruction on all $24$ Kodak images at native resolution, three
seeds, every activation tuned to its best set of internal parameters. Members of the proposed family of activations are shown in italics.
A dash (``--'') marks a quantity that is undefined for an activation with unbounded
support. The symbol ``?'' marks a run that did not reach $30$~dB within the 5000 iterations training budget.
Effective rank is out of a maximum of $256$.}
\label{tab:native}
\vspace{-5pt}
{\providecommand{\tabsep}{5pt}\setlength{\tabcolsep}{\tabsep}
\begin{tabular*}{\linewidth}{@{\extracolsep{\fill}}lccccc@{}}
\toprule
Activation & PSNR (dB) & SSIM & inactive (\%) & effective rank & iters to 30 dB \\
\midrule
SIREN & 38.55\,\tiny{$\pm$2.86} & 0.956 & --- & 241 & 2300 \\
FINER & 38.28\,\tiny{$\pm$2.68} & 0.955 & --- & 244 & 1700 \\
\textit{JaSIN, $n{=}4$} & 37.28\,\tiny{$\pm$2.99} & 0.947 & 1.5 & 238 & 2400 \\
WIRE (Gabor) & 35.75\,\tiny{$\pm$2.77} & 0.934 & --- & 227 & 3100 \\
\textit{JaSIN, $n{=}2$} & 34.59\,\tiny{$\pm$3.01} & 0.923 & 1.8 & 226 & 3200 \\
ReLU+RFF & 33.36\,\tiny{$\pm$3.03} & 0.884 & 16.1 & 142 & 1900 \\
Gaussian & 31.62\,\tiny{$\pm$3.31} & 0.861 & --- & 180 & 3800 \\
\textit{JaSIN, $n{=}1$} & 30.97\,\tiny{$\pm$3.52} & 0.835 & 4.5 & 184 & 4200 \\
Triangle, 1-sided & 27.33\,\tiny{$\pm$3.16} & 0.708 & 57.8 & 79 & ? \\
ReLU & 23.98\,\tiny{$\pm$2.98} & 0.557 & 41.6 & 39 & ? \\
Tanh & 23.00\,\tiny{$\pm$2.63} & 0.513 & --- & 42 & ? \\
\bottomrule
\end{tabular*}
}
\end{table}

\vfill
\paragraph{Repair} Table~\ref{tab:resurrect} reports the effect of reinitializing
inactive units during training. Both rules eliminate inactivity entirely, from
as much as $68\%$ down to zero. Their effect on reconstruction depends on the
regime. At $b=0.1$, where inactivity is mild, centering on the origin is best at
every oscillation count, by $0.35$~dB at $n\!=\!1$ but only
$0.02$--$0.12$~dB beyond it; the linear piece it targets has width $2b/n$. At $b=0.02$, where
the support is too narrow for that piece to retain a unit, naive
reinitialization is better, and the gain over no intervention is large, reaching
$7.6$~dB at $n\!=\!4$. At $b=0.5$ the support is wide enough that no unit is
certified inactive beyond $n\!=\!1$, and both rules are then exact no-ops: all
three columns agree to single-precision round-off. Reconstruction is nonetheless
$5.0$~dB below the $b=0.1$ optimum at $n\!=\!4$, so driving inactivity to zero
is not itself the objective.

\begin{table}[t]
\centering\scriptsize
\caption{Reinitializing provably inactive units every $K$ iterations.}
\label{tab:resurrect}
\vspace{-5pt}
{\providecommand{\tabsep}{3pt}\setlength{\tabcolsep}{\tabsep}
\begin{tabular*}{\linewidth}{@{\extracolsep{\fill}}cccccccc@{}}
\toprule
& & \multicolumn{3}{c}{PSNR (dB)} & \multicolumn{3}{c}{inactive (\%)} \\
\cmidrule(lr){3-5}\cmidrule(lr){6-8}
$b$ & $n$ & none & random & centered & none & random & centered \\
\midrule
$0.02$ & 1 & 25.46 & \textbf{25.62} & 23.44 & 68.2 & 0.0 & 0.0 \\
 & 2 & 29.13 & \textbf{33.18} & 33.00 & 36.9 & 0.0 & 0.0 \\
 & 4 & 26.62 & \textbf{34.26} & 32.67 & 26.9 & 0.0 & 0.0 \\
\midrule
$0.1$ & 1 & 30.97 & 31.30 & \textbf{31.66} & 4.5 & 0.0 & 0.0 \\
 & 2 & 34.59 & 34.59 & \textbf{34.61} & 1.8 & 0.0 & 0.0 \\
 & 4 & 37.28 & 37.32 & \textbf{37.44} & 1.5 & 0.0 & 0.0 \\
\midrule
$0.5$ & 1 & 27.08 & 27.32 & \textbf{27.52} & 0.9 & 0.0 & 0.0 \\
 & 2 & 30.19 & 30.19 & 30.19 & 0.0 & 0.0 & 0.0 \\
 & 4 & 32.29 & 32.29 & 32.29 & 0.0 & 0.0 & 0.0 \\
\bottomrule
\end{tabular*}
}

\end{table}

\newpage

\paragraph{Compression} Table~\ref{tab:prune} removes every provably inactive
unit and rebuilds the network with narrower layers. The reconstruction is
identical to single-precision round-off in every case, while up to
$\prunemaxparams\%$ of parameters are eliminated and decoding is accelerated by
up to $\prunemaxspeed\times$. The magnitude baseline removes from each layer
the same number of units the certificate removes, choosing those of smallest
$\ell_1$ outgoing weight norm and likewise without retraining, as is standard
for structured pruning~\cite{han2015learning,li2017pruning}; it costs between
$\prunecostmin$ and $\prunecostmax$~dB, and the severity is not predictable from the sparsity
level: magnitude pruning of $\pruneworstunits\%$ of the units of $\jasin_4$
costs $\prunecostmax$~dB, whereas removing $\prunebestunits\%$ of those of the
one-sided triangle costs $\prunecostmin$~dB.

\begin{table}[t]
\centering\scriptsize
\caption{Removing every provably inactive unit, and the same number per layer
chosen by smallest outgoing weight norm, at native resolution. Activations
with no inactive units are omitted.}
\label{tab:prune}
\vspace{-5pt}
{\providecommand{\tabsep}{3pt}\setlength{\tabcolsep}{\tabsep}
\begin{tabular*}{\linewidth}{@{\extracolsep{\fill}}lccccc@{}}
\toprule
& \multicolumn{2}{c}{PSNR (dB)} & \multicolumn{3}{c}{removed} \\
\cmidrule(lr){2-3}\cmidrule(lr){4-6}
Activation & exact & magnitude & units (\%) & parameters (\%) & speedup \\
\midrule
Triangle, 1-sided & \textbf{27.33} & 26.24 & 57.8 & 83.6 & 3.22$\times$ \\
ReLU & \textbf{23.98} & 21.29 & 41.6 & 70.3 & 2.03$\times$ \\
ReLU+RFF & \textbf{33.36} & 21.81 & 16.1 & 20.4 & 1.11$\times$ \\
\textit{JaSIN, $n{=}1$} & \textbf{30.97} & 28.51 & 4.5 & 6.7 & 1.06$\times$ \\
\textit{JaSIN, $n{=}2$} & \textbf{34.59} & 31.48 & 1.8 & 2.2 & 1.01$\times$ \\
\textit{JaSIN, $n{=}4$} & \textbf{37.28} & 25.35 & 1.5 & 1.8 & 1.00$\times$ \\
\bottomrule
\end{tabular*}
}

\end{table}

\paragraph{Cost} Using no transcendental evaluation, $\jasin$ might be
expected to train faster than the activations it is compared against. On a GPU
it does not: an activation's cost is governed by traffic between
processor and memory rather than by its arithmetic, and a sine is a single
hardware operation, so the whole sweep spans $7.1$--$8.6$~ms per step and
$\jasin_4$ sits within $1\%$ of SIREN. The piecewise-linear saving appears only
where the arithmetic is the bottleneck, as on a single CPU thread, a regime that matters on edge
hardware~\cite{dupont2021coin}, for example. There the
activation alone is $\cpuspeedup\times$ faster than a sine and
$\cpuspeedupwire\times$ faster than a Gabor wavelet, within $1.2\times$ of ReLU

\subsection{Discussion}
\label{sec:disc}
These results identify two properties ReLU lacks that help INR image
fitting. The first is a two-sided compact support centered on the origin:
replacing the clipped ReLU with such a bump ($\jasin$, $n\!=\!1$), which is the
one-bump and the centering step of Table~\ref{tab:ladder} taken together,
contributes $\laddersupport$~dB. The second is oscillation: raising the
oscillation count to $n\!=\!4$ contributes another $\ladderosc$~dB. Clipping
ReLU alone contributes $\ladderbounded$~dB. Both properties can be supplied without leaving the piecewise-linear
class. The oscillation count can then be seen as a
capacity parameter~\cite{telgarsky2016benefits},
where composing a map with $n$ oscillations multiplies the piece count by $n$ at
every layer.

It is not, however, a free improvement. Beyond $n\!=\!4$ the reconstruction
collapses to $15$~dB while the effective rank reaches its highest value, $248$
of $256$: the layers use nearly every direction available to them and the fit
degrades anyway, so the sweep measures a combination of capacity and
trainability rather than capacity alone. Restricting the support is not the limiting factor in the range
we examine, since the compactly supported $n\!=\!4$ activation outperforms the
unbounded periodic triangle by $2.8$~dB, but a true sinusoid outperforms its
piecewise-linear counterpart (the periodic triangular bump) by $4$~dB and we cannot attribute that difference
to the support.

\begin{figure}[t]
\centering
\begin{subfigure}{0.242\linewidth}\centering
  \includegraphics[width=\linewidth]{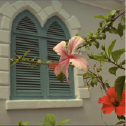}
  \caption{support $\whichsupa$\\oscillation\ $\whichosca$}\end{subfigure}\hfill
\begin{subfigure}{0.242\linewidth}\centering
  \includegraphics[width=\linewidth]{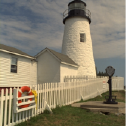}
  \caption{support $\whichsupb$\\oscillation\ $\whichoscb$}\end{subfigure}\hfill
\begin{subfigure}{0.242\linewidth}\centering
  \includegraphics[width=\linewidth]{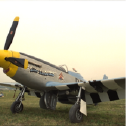}
  \caption{support $\whichsupc$\\oscillation\ $\whichoscc$}\end{subfigure}\hfill
\begin{subfigure}{0.242\linewidth}\centering
  \includegraphics[width=\linewidth]{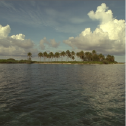}
  \caption{support $\whichsupd$\\oscillation\ $\whichoscd$}\end{subfigure}
\vspace{-5pt}
\caption{Performance gains on 4 tested images.
``support'' is the gain in dB from replacing the one-sided bump with a compact
two-sided one (i.e. clipped ReLU to $\jasin_1$); ``oscillation'' is the further gain from
raising the oscillation count ($n$) from one to four.
}
\label{fig:which}
\end{figure}

The two contributions are approximately equal in the mean across the dataset but not image by image. Completing
the experimental design with a one-sided activation at $n\!=\!4$ makes the interaction
 between the two properties above estimable: oscillation is worth $5.57$~dB without two-sided support and
$6.30$~dB with it, on average, so
there is no evidence that the two properties substitute for one another. What
varies is the split: the total gain has a standard deviation of $1.5$~dB across
images, while the support term ranges from $4.2$ to $13.1$~dB and the
oscillation term from $3.9$ to $9.0$. Figure~\ref{fig:which} suggests the split follows content: the
image gaining most from support is built from flat regions and hard edges, the
one gaining most from oscillation from fine, smooth texture.

Bounded support also changes what can be known about a trained network. The
inactivity test is exact rather than heuristic in this setting, which converts
two operations that are ordinarily approximate, namely repairing inactive units
during training and removing them afterwards, into exact ones.

Finally, three qualifications apply. As shown in Table \ref{tab:native}, the family does not reach the highest
PSNR; so here the contribution is the diagnosis. It is not uniformly cheaper either,
since the piecewise-linear arithmetic pays only where arithmetic rather than
memory traffic is the bottleneck, which on a GPU it is not.
Finally, the inactivity test is exact only on the training set. For an INR, that
is the entire domain, which is what makes the pruning guarantee complete, but
the argument does not transfer to tasks with unseen test inputs.

\vfill
\section{Conclusion}
\label{sec:conc}
What ReLU lacks for implicit neural representations is two properties: compact two-sided support and oscillation, worth $\laddersupport$ and
$\ladderosc$~dB and both supplied without leaving the piecewise-linear class.
Together they raise ReLU to within $1.27$~dB of a tuned, state-of-the-art sinusoid. Compact
support additionally makes inactivity decidable, which a globally supported
activation cannot. This proves that capacity is unused and, for a representation fitted to its own evaluation grid, asserts exact retraining-free compression.

For future work, we will measure what properties cost on tasks ReLU already performs well, such as image
classification and object detection, and study the use of $\jasin$ as potential bridge in the understanding of these tasks and Implicit Neural Representations.

\vfill\pagebreak
\bibliographystyle{IEEEbib}
\bibliography{refs}
\end{document}